# Determining the Unithood of Word Sequences using Mutual Information and Independence Measure

**Wilson Wong, Wei Liu and Mohammed Bennamoun**
School of Computer Science and Software Engineering
University of Western Australia
Crawley WA 6009
{wilson,wei,bennamou}@csse.uwa.edu.au

## Abstract

Most works related to unithood were conducted as part of a larger effort for the determination of termhood. Consequently, the number of independent research that study the notion of unithood and produce dedicated techniques for measuring unithood is extremely small. We propose a new approach, independent of any influences of termhood, that provides dedicated measures to gather linguistic evidence from parsed text and statistical evidence from Google search engine for the measurement of unithood. Our evaluations revealed a precision and recall of 98.68% and 91.82% respectively with an accuracy at 95.42% in measuring the unithood of 1005 test cases.

## 1 Introduction

Terms and the tasks related to their treatments are an integral part of many applications that deal with natural language text such as large-scale search engines, automatic thesaurus construction, machine translation and ontology learning for purposes ranging from indexing to cluster analysis. With the increasing reliance on huge text sources such as the World Wide Web as input, the need to provide automated means for managing domain-specific terms rises. Such relevance and importance of terms has prompted dedicated research interests. Various names such as *automatic term recognition*, *term extraction* and *terminology mining* were given to encompass the tasks related to the treatment of terms. Term extraction is the process of extracting lexical units from text and filtering them for the purpose of identifying terms which characterise certain domains of interest. This process is the determination of two important factors, namely, *unithood* and *termhood*. Unithood concerns with whether or not a sequence of words should be combined to form a more stable lexical unit, and termhood measures the degree to which these stable lexical units are related to domain-specific concepts. While unithood is only relevant to *complex terms* (i.e. multi-word terms), termhood concerns both *simple terms* (i.e. single-word terms) and complex terms.

Most research in automatic term recognition were conducted solely to study and develop techniques for measuring termhood, while only a small number exists that study on unithood. Unfortunately, rather than considering the measurement of unithood as an important prerequisite, these researchers merely treat it as part of a larger scoring and filtering mechanism for determining termhood. Such perception is clearly reflected through the words of Kit (2002), *"...we can see that the unithood is actually subsumed, in general, by the termhood."*. Consequently, the significance of unithood measures has been overshadowed by the larger notion of termhood. As such, the progress and innovation with respect to this small sub-field of automatic term recognition is minimal. Most of the existing techniques for measuring unithood employ conventional measures such as mutual information and log-likelihood, and rely simply on the occurrence and co-occurrence frequencies from domain corpora as the source of evidence.

In this paper, we propose the separation of unithood measurements from the determination of termhood. From here on, we will consider unithood measurement as an important prerequisite, rather than a subsumption, to the determination of termhood. We present a new dedicated approach for determining the unithood of word sequences by employing the Google search engine as the source of statistical evidence, and measures inspired by mutual information (Church and Hanks (1990)) and *Cvalue* (Frantzi (1997)). The use of the World Wide Web to replace the conven-tional use of static corpora will eliminate issues related to portability to other domains, and the size of text necessary to induce the required statistical evidence. Besides, this dedicated approach to determine the unithood of word sequences will prove to be invaluable to other areas in natural language processing such as noun-phrase chunking and named-entity recognition. Our evaluations revealed a precision and a recall of 98.68% and 91.82% respectively with an accuracy at 95.42% in measuring the unithood of 1005 test cases.

In Section 2, we have a brief review on the existing techniques for measuring unithood. In Section 3, we present our new approach, the measures involved and the justification behind every aspect of the measures. In Section 4, we summarize some findings from our evaluations. We discuss in Section 5 why our new approach can be applicable to other tasks in natural language processing such as named-entity recognition. Finally, we conclude this paper with an outlook to future works in Section 6.

## 2 Related Works

Prior to measuring unithood, term candidates must be extracted. There are two common approaches for extracting the term candidates. The first requires the corpus to be tagged or parsed, and a filter is then employed to extract words or phrases satisfying some linguistic patterns. There are two types of filters for extracting from tagged corpus, namely, open or closed. Too restricted filters (i.e. closed) that rely on a small set of allowable part-of-speech will produce high precision but poor recall (Frantzi and Ananiadou (1997)). On the other hand, filters that are too liberal (i.e. open), allowing part-of-speech such as prepositions and adjectives, will have the opposite effect. Most of the existing approaches rely on regular expressions and the part-of-speech tags to accept or reject sequences of n-grams as term candidates. For example, Frantzi and Ananiadou (1997) employ Brill tagger to tag the raw corpus with part-of-speech and later extract n-grams that fulfill the pattern $(Noun|Adjective)^+Noun$. Bourigault and Jacquemin (1999) utilise SYLEX, a part-of-speech tagger, to tag the raw corpus. The part-of-speech tags are utilised to extract maximal-length noun phrases, which are later recursively decomposed into heads and modifiers. On the other extreme, Dagan and Church (1994) accept only sequences of $Noun+$. The second type of extraction approaches works on raw corpus using a set of heuristics. This type of approaches which does not rely on part-of-speech tags is quite rare. Such approaches have to make use of the textual surface constraints to approximate the boundaries of term candidates. One of the constraints include the use of a stopword list to obtain the boundaries of stopwords for inferring the boundaries of candidates. A selection list of allowable prepositions can also be employed to enforce constraints on the tokens between units.

The filters for extracting term candidates make use of only local, surface-level information, namely, the part-of-speech tags. More evidence is required to establish the dependence between the constituents of each term candidate to ensure strong unithood. Such evidence will usually be statistical in nature in the form of co-occurrences of the constituents in the corpus. Accordingly, the unithood of the term candidates can be determined either as a separate step or may proceed as part of the extraction process. From our review of the literature, only an extremely small number of researchers actually discussed and presented measures for unithood. The lack of extensive research and techniques related specifically to unithood is reaffirmed by Kit (2002). According to the author, *"...its measure (if there is one) indicates how likely it is that a term candidate is an atomic text unit."*

Two of the most common measures of unithood have to be pointwise *mutual information (MI)* (Church and Hanks (1990)) and *log-likelihood ratio* (Dunning (1994)). In mutual information, the co-occurrence frequencies of the constituents of complex terms are utilised to measure their dependency. The mutual information for two words *a* and *b* is defined as:

$$MI(a,b) = \log_2 \frac{p(a,b)}{p(a)p(b)} \quad (1)$$

where $p(a)$ and $p(b)$ are the probabilities of occurrence of *a* and *b*. Many measures that apply statistical techniques assuming strict normal distribution, and independence between the word occurrences do not fare well. For handling extremely uncommon words or small sized corpus, *log-likelihood ratio* delivers the best precision (Kurz and Xu (2002); Franz (1997)). Log-likelihood ratio attempts to quantify how much more likely one pair of words is to occur compared to the others. Despite its potential, *"How to apply*

*this statistic measure to quantify structural dependency of a word sequence remains an interesting issue to explore."* (Kit (2002)).

Frantzi (1997) proposed a measure known as *Cvalue* for extracting complex terms. The measure is based upon the claim that a substring of a term candidate is a candidate itself given that it demonstrates adequate independence from the longer version it appears in. For example, *"E. coli food poisoning"*, *"E. coli"* and *"food poisoning"* are acceptable as valid complex term candidates. However, *"E. coli food"* is not. Therefore, some measures are required to gauge the strength of word combinations to decide whether two word sequences should be merged or not. Given a word sequence *a* to be examined for unithood, the *Cvalue* is defined as:

$$Cvalue(a) = \begin{cases} \log_2 |a| \cdot f_a & \text{if } |a| = g \\ \log_2 |a| \cdot (f_a - \frac{\sum_{l \in L_a} f_l}{|L_a|}) & \text{otherwise} \end{cases}$$
(2)

where $|a|$ is the number of words in $a$, $L_a$ is the set of longer term candidates that contain $a$, $g$ is the longest n-gram considered, $f_a$ is the frequency of occurrence of $a$, and $a \notin L_a$. While certain researchers (Kit (2002)) consider *Cvalue* as a termhood measure, others (Nakagawa and Mori (2002)) accept it as a measure for unithood. One can observe that longer candidates tend to gain higher weights due to the inclusion of $log_2|a|$ in Equation 2. In addition, the weights computed using Equation 2 are purely dependent on the frequency of $a$.

## 3 A New Approach for Unithood Measurement

Our new approach for measuring the unithood of word sequences consists of two parts. Firstly, a list of word sequences is extracted using purely linguistic techniques. Secondly, word sequences are examined and the related statistical evidence is gathered to assist in determining their mutual information and independence.

### 3.1 Extracting Word Sequences

Existing techniques for extracting word sequences have been relying on part-of-speech information and filters in the form of pattern matching (e.g. regular expression). Since the head-modifiers principle is important for our techniques, we employ both the part-of-speech information and dependency relation for extracting term candidates. The filter is implemented as a head-driven left-right filter (Wong (2005)) that feeds on the output of Stanford Parser (Klein and Manning (2003)), which is an implementation of unlexicalised *probabilistic context-free grammar (PCFG)* and lexical dependency parser. The head-driven filter begins by identifying a list of head nouns from the output of the Stanford Parser. As the name suggests, the filter begins from the head and proceeds to the left and later, right in the attempt to identify maximal-length noun phrases according to the head-modifier information. During the process, the filter will append or prepend any immediate modifier of the current head which is a noun (except possessive nouns), an adjective or a foreign word. Each noun phrase or segment of noun phrase identified using the head-driven filter is known as a *potential term candidate*, $a_i \in A$ where $i$ is the word offset produced by the Stanford Parser (i.e. the *"offset"* column in Figure 1).

Figure 1 shows the output of the Stanford Parser for the sentence *"They're living longer with HIV in the brain, explains Kathy Kopnisky of the NIH's National Institute of Mental Health, which is spending about millions investigating neuroAIDS."*. Note that the words are lemmatised to obtain the root form. The head nouns are marked with squares in the figure. For example, the head *"Institute"* is modified by *"NIH's"*, *"National"* and *"of"*. Since we do not allow for modifiers of the type *"possessive"* and *"preposition"*, we will obtain *"National Institute"* as shown in Figure 2. Figure 2 shows the head-driven filter at work for some of the head nouns identified from the *"modifiee"* column of the output in Figure 1. After the head-driven filter has identified potential term candidates using the heads, remaining nouns from the *"word"* column in Figure 1 which are not part of any potential term candidates will be included in $A$.

### 3.2 Determining the Unithood of Word Sequences

In the following step, we examine the unithood of all pairs of potential term candidates $(a_x, a_y) \in A$ with $a_x$ and $a_y$ located immediately next to each other (i.e. $x+1 = y$), or separated by a preposition or coordinating conjunction *"and"* (i.e. $x+2 = y$). Obviously, $a_x$ has to appear before $a_y$ in the sentence or in other words, $x < y$ for all pairs where $x$ and $y$ are the word offsets produced by the Stan-ford Parser. Formally, given that $s = a_x b a_y$ where $b$ is any preposition, the conjunction *"and"* or an empty string, the problem is to determine whether to accept *s* as an independent lexical unit (i.e. a term candidate) or leave $a_x$ and $a_y$ as separate units. In order to decide on the merge, we need adequate evidence that *s* will form a stable unit and hence, a better term candidate than $a_x$ and $a_y$ separated. It is worth mentioning that the size (i.e. number of words) of $a_x$ and $a_y$ is not limited to 1. For example, we can have $a_x$=*"National Institutes"*, $b$=*"of"* and $a_y$=*"Allergy and Infectious Diseases"*. In addition, the size of $a_x$ and $a_y$ should have no effect on the determination of their unithood.

| offset | word | partofspeech | dependency | modifiee | modifiee offset |
|---|---|---|---|---|---|
| 2 | They | PRP | nsubj | live | 4 |
| 3 | are | VBP | aux | live | 4 |
| 4 | live | VBG | ccomp | explains | 13 |
| 6 | with | IN | dep | long | 5 |
| 7 | HIV | NN | pobj | with | 6 |
| 8 | in | IN | prep | live | 4 |
| 9 | the | DT | det | brain | 10 |
| 10 | brain | NN | pobj | in | 8 |
| 14 | Kathy | NNP | nn | Kopnisky | 15 |
| 15 | Kopnisky | NNP | nsubj | explains | 13 |
| 16 | of | IN | prep | Kopnisky | 15 |
| 17 | the | DT | det | NIH's | 18 |
| 18 | NIH's | NNP | poss | Institute | 21 |
| 20 | National | NNP | nn | Institute | 21 |
| 21 | Institute | NNP | pobj | of | 16 |
| 22 | of | IN | prep | Institute | 21 |
| 23 | Mental | NNP | nn | Health | 24 |
| 24 | Health | NNP | pobj | of | 22 |
| 26 | which | WDT | rel | spend | 28 |
| 27 | is | VBZ | aux | spend | 28 |
| 28 | spend | VBG | rcmod | Kopnisky | 15 |
| 30 | million | CD | dobj | spend | 28 |
| 31 | investigate | VBG | partmod | spend | 28 |
| 32 | neuroAIDS | NNS | dobj | investigate | 31 |

Figure 1: The output from Stanford Parser. The tokens in the *"modifiee"* column marked with squares are head nouns, and the corresponding tokens along the same rows in the *"word"* column are the modifiers. The first column *"offset"* is subsequently represented using the variable *i*.

The most suitable existing measure for gathering evidence about the dependency between two words is mutual information. Based on the conventional practice, the frequency of occurrence of each element in $W = \{s, a_x, a_y\}$ is normalized using the sum of the frequency of all occurrences in $W$. Since data sparseness in the local corpus may lead to poor estimation of mutual information, we innovatively employ the page count by Google search engine for calculating the dependency of the elements in $W$ instead. We treat the World Wide Web as a large general corpus and Google search engine as a gateway for accessing the documents in the corpus. Our choice of using Google to obtain the page count was merely motivated by its extensive coverage. In fact, it is possible to employ any search engines on the World Wide Web for this research. Each element in $W$ is formulated as a search query and submitted to Google search engine. The page count returned is utilised for calculating the mutual information. In addition, we also apply a multiplier within the range of $[e^{-1}, 1]$, inspired by TF-IDF, to offset too common terms, especially in three-word terms such as *"Institute of Science"*. Formally, for each $w \in W$, we define the weight as:

$$p(w) = \frac{n_w}{\sum_{v \in W} n_v} e^{\left(-\frac{n_w}{\sum_{v \in W} n_v}\right)} \quad (3)$$

where $n_w$ is the page count (i.e. number of documents) returned by Google search engine containing $w \in W$. We only take into consideration the number of documents that contain the word sequences (i.e. page count) due to the difficulty in obtaining the actual frequency of occurrences of word sequences from Google's search results. Next, the mutual information between the two units $a_x$ and $a_y$ is defined as:

```
<< (START-OF-HASH) << They(2) << are(3) << live(4) << with(6) << HIV(7) << in(8) << the(9)
<< brain(10) << Kathy(14) << Kopnisky(15) << of(16) << the(17) << NIH(18) << National(20)
<< Institute(21) << of(22) << Mental(23) << Health(24) >> which(26) >> is(27) >> spend(28)
>> million(30) >> investigate(31) >> neuroAIDS(32) >> (END-OF-HASH) >>

<< (START-OF-HASH) << They(2) << are(3) << live(4) << with(6) << HIV(7) << in(8) << the(9)
<< brain(10) << Kathy(14) << Kopnisky(15) << of(16) << the(17) << NIH(18) << National(20)
<< Institute(21) >> of(22) >> Mental(23) >> Health(24) >> which(26) >> is(27) >> spend(28)
>> million(30) >> investigate(31) >> neuroAIDS(32) >> (END-OF-HASH) >>

<< (START-OF-HASH) << They(2) << are(3) << live(4) << with(6) << HIV(7) << in(8) << the(9)
<< brain(10) << Kathy(14) << Kopnisky(15) >> of(16) >> the(17) >> NIH(18) >> National(20)
>> Institute(21) >> of(22) >> Mental(23) >> Health(24) >> which(26) >> is(27) >> spend(28)
>> million(30) >> investigate(31) >> neuroAIDS(32) >> (END-OF-HASH) >>
```

Figure 2: An example of our head-driven left-right filter at work. The tokens which are highlighted with a darker tone are the head nouns. The underlined tokens are the modifiers identified as a result of the left-first and right-later movement of the filter. In the first segment, the head noun is *"Health"* while the first token to the left *"Mental"* is the corresponding modifier. In the second segment of the figure, *"Institute"* is the head noun and *"National"* is the modifier. Note that the results from the first and second segments are actually part of a longer noun phrase. Due to the restriction in accepting prepositions by the head-driven filter, *"of"* is omitted from the output of the first segment of the figure.

$$MI(a_x, a_y) = \frac{p(s)}{p(a_x)p(a_y)} \quad (4)$$

If the occurrence of *s* approaches 0 (i.e. $a_x$ and $a_y$ are rarely seen together as a unit with *b*), then $MI(a_x, a_y)$ reduces to 0. If the occurrence of *s* approaches that of the co-occurrence of $a_x$ and $a_y$, $MI(a_x, a_y)$ will approach 1. A high $p(s)$ indicates stronger coupling of $a_x$ and $a_y$, or even more so, the existence of $a_x$ and $a_y$ are purely due to the existence of *s*. A high $MI(a_x, a_y)$ implies an increase in the unithood of the two units $a_x$ and $a_y$. Following this, $a_x$ and $a_y$ will have poor unithood in their individual forms if they are not merged into *s* to form a stronger unit. This mutual information measure is necessary in distinguishing phrases such as *"Asia and Europe"* (i.e. low mutual information) from *"U.S. Food and Drug Administration"* (i.e. high mutual information) when prepositions and conjunctions are involved.

Nonetheless, the units $a_x$ and $a_y$ may still be capable of forming valid compound unit *s* even though their mutual information is relatively low. Low mutual information can be attributed to the high individual occurrences of $a_x$ and $a_y$ due to their extremely common usage. For example, $a_x$=*"Institute"* and $a_y$=*"Ophthalmology"* yield high occurrences relative to *s*=*"Institute of Ophthalmology"*. This does not mean that *"Institute of Ophthalmology"* is not a valid unit. To handle such cases where $MI(a_x, a_y)$ is mediocre due to the commonness of $a_x$ and $a_y$, we employ another measure of independence. In such situation, we will still accept *s* as a valid unit if it can be demonstrated that the extremely high independence of the individual unit $a_x$ and $a_y$ is the cause behind the low $MI(a_x, a_y)$. For this purpose, we modify the *Cvalue* described in Equation 2 to accommodate the use of page counts rather than frequency. In addition, we remove the multiplier $\log_2 |a|$ because the number of words in $a_x$ and $a_y$ does not play a role in determining their independence from *s*. Consequently, we define the measure of *Independence (ID)* for $a_x$ and $a_y$ from *s* as:

$$ID(a_x, s) = \begin{cases} \log_{10}(n_{a_x} - n_s) & \text{if}(n_{a_x} > n_s) \\ 0 & \text{otherwise} \end{cases} \quad (5)$$

$$ID(a_y, s) = \begin{cases} \log_{10}(n_{a_y} - n_s) & \text{if}(n_{a_y} > n_s) \\ 0 & \text{otherwise} \end{cases} \quad (6)$$

where $n_{a_x}$, $n_{a_y}$ and $n_s$ is the Google page count for the unit $a_x$, $a_y$ and *s*, respectively. As the lexical unit $a_x$ occurs more than its longer counterpart *s*, its independence $ID(a_x, s)$ grows. Only when the number of occurrences of $a_x$ is less than those of *s*, its independence from *s* becomes $ID(a_x, s) = 0$. This means that we will not be able to witness $a_x$ without encountering *s*. The same can be said about the measure of independence for $a_y$, $ID(a_y, s)$. In short, extremely high independence of $a_x$ and $a_y$ relative to *s* will be reflected through high $ID(a_x, s)$ and $ID(a_y, s)$.

Consequently, the decision to merge $a_x$ and $a_y$ to form $s$ depends on both the mutual information between $a_x$ and $a_y$, namely, $MI(a_x, a_y)$, and the independence of $a_x$ and $a_y$ from $s$, namely, $ID(a_x, s)$ and $ID(a_y, s)$. This decision is organised into a Boolean function known as *Unithood* ($UH$), and we define it as:

$$UH(a_x, a_y) = \begin{cases} 1 & \text{if } (MI(a_x, a_y) > MI^+) \vee \\ & (MI^+ \geq MI(a_x, a_y) \\ & \geq MI^-  \wedge \\ & ID(a_x, s) \geq ID_T \wedge \\ & ID(a_y, s) \geq ID_T \wedge \\ & IDR^+ \geq IDR(a_x, a_y) \\ & \geq IDR^-) \\ 0 & \text{otherwise} \end{cases}$$
(7)

where $IDR(a_x, a_y) = ID(a_x, s)/ID(a_y, s)$. $IDR$ helps to ensure that pairs with mediocre mutual information not only have $a_x$ and $a_y$ with high independence but are also equally independent before mergings are performed. The unithood function in Equation 7 summarises the relationship between mutual information and the independence measure. $UH(a_x, a_y)$ simply states that the two lexical units $a_x$ and $a_y$ can only be merged in two cases:

- If $a_x$ and $a_y$ has extremely high mutual information (i.e. higher than a certain threshold $MI^+$); or

- If $a_x$ and $a_y$ achieve average mutual information (i.e. within the acceptable range of two thresholds $MI^+$ and $MI^-$) due to both of their extremely high independence (i.e. higher than the threshold $ID_T$) from $s$. To ensure that both units have equally high independence, their ratio of independence $IDR$ has to fall within the range $IDR^-$ and $IDR^+$.

The thresholds for $MI(a_x, a_y)$, $ID(a_x, s)$, $ID(a_x, s)$ and $IDR(a_x, a_y)$ are decided empirically through our evaluations:

- $MI^+ = 0.9$
- $MI^- = 0.02$
- $ID_T = 6$
- $IDR^+ = 1.35$
- $IDR^- = 0.93$

A general guideline for determining the appropriate threshold values is provided in the next section. Finally, the word sequence $s = a_x b a_y$ will be accepted as a stable lexical unit (i.e. term candidate) if and only if $UH(a_x, a_y) = 1$.

## 4 Evaluations and Discussions

Table 1: Contingency table constructed using the actual and ideal results for computing precision, recall, accuracy and F-score. *Ideal results* are used as a reference for evaluation. *Actual results* are the actual output from our new approach.

| contingency table | | ideal results | | |
| --- | --- | --- | --- | --- |
| | | merged | not merged | |
| actual results | merged | 449 | 6 | 455 |
| | not merged | 40 | 510 | 550 |
| | | 489 | 516 | 1005 |

For this evaluation, we employ 300 news articles from Reuters in the health domain gathered between October 2006 to January 2007. These 300 articles are fed into the Stanford Parser whose output is then used by our head-driven left-right filter to extract word sequences in the form of nouns and noun phrases. Pairs of word sequences (i.e. $a_x$ and $a_y$) located immediately next to each other, or separated by a preposition or the conjunction *"and"* in the same sentence are measured for their unithood. Based on the $UH(a_x, a_y)$ of the pairs, the decisions on whether to merge or not are done automatically. These decisions are known as the *actual results*. At the same time, we inspect the same list manually to decide on the merging of all the pairs. These decisions are known as the *ideal results*. Using the 300 news articles, we managed to obtain 1005 pairs of words to be tested for unithood. The actual and ideal results are organised into a contingency table as shown in Table 1 to identify the true and the false positives, and the true and the false negatives. Using the results in Table 1, we obtained a precision of 98.68%, a recall of 91.82% and an F-score of 90.61%. As for the accuracy, our new measures for unithood scored 95.42%. It shows that our new measure has very good precision and a relatively low recall due to the high number of false negatives.

Firstly, we realised that the high false negative rate is explained by our more conservative definition of the thresholds namely $ID_T$, $IDR^+$ and $IDR^-$. We discovered that about 90% of the false negatives fall within the range of $MI^+$ and $MI^-$ (i.e. mediocre mutual information). Such pairs have the opportunity to be merged if they demonstrate adequate independence from s. Unfortunately, most of the independence ID of either one or both members of the pairs failed to satisfy our independence thresholds. For example, referring to rows 72, 120 and 567 as shown in Figure 3, one will notice that they have mediocre mutual information as defined by $MI^+ = 0.9$ and $MI^- = 0.02$. In the case of row 72, both $a_x$ and $a_y$ have an independence lower than the threshold $ID_T = 6$. This resulted in the decision of not merging them. The same case happened to row 120. In row 567, only $a_y$ has an independence lower than $ID_T$ and at the same time, their IDR is well above the upper limit $IDR^+$. The remaining 10% of the false negatives are simply due to the extremely low mutual information $MI(a_x, a_y)$ of the pairs. Take for example pair 171 in Figure 3 where the mutual information is only 0.002, which is way below $MI^-$. Secondly, due to the small number of false positives, not much conclusion can be drawn. From our analysis of the results, most of the false positives are due to their high mutual information (i.e. $MI(a_x, a_y)$ above $MI^+$). Pair 66 in Figure 3 is an example which is incorrectly merged due to a mutual information value higher than $MI^+$.

| no. | $a_x$ | b | $a_y$ | $ID(a_x,s)$ | $ID(a_y,s)$ | $IDR(a_x,a_y)$ | $MI(a_x,a_y)$ | $UH(a_x,a_y)$ | s |
|---|---|---|---|---|---|---|---|---|---|
| 66 | Agriculture Department lab | in | Ames | 1.6989 | 7.3765 | 0.2303 | 1.5466 | MERGED | Agriculture Department lab in Ames |
| 72 | EU Commissioner | for | Health and Consumer Protection Markos Kyprianou | 5.9029 | 2.7481 | 2.1479 | 0.7289 | NOTMERGED | EU Commissioner for Health and Consumer Protection Markos Kyprianou |
| 120 | Atlanta-based Centers | for | Disease Control and Prevention | 4.0583 | 5.9370 | 0.6835 | 0.1932 | NOTMERGED | Atlanta-based Centers for Disease Control and Prevention |
| 171 | Joint Committee | on | Public Health | 6.0488 | 8.2041 | 0.7372 | 0.0021 | NOTMERGED | Joint Committee on Public Health |
| 567 | Center | for | Tobacco Research | 8.9216 | 5.6280 | 1.5852 | 0.1529 | NOTMERGED | Center for Tobacco Research |

Figure 3: This figure shows a snapshot of some samples of false positives and false negatives taken from our evaluations. Row 66 is an example of merged pair which is not supposed to be combined (i.e. false positives) while the remaining rows are false negatives. Each row has two lexical units $a_x$ (column 2) and $a_y$ (column 4) to be examined for their mutual information (column 8), independence from s (column 5 and 6) to determine if they should be merged into s (column 10). The decision to merge (column 9) is accomplished based on Equation 7.

From our discussion above, one would realise that both the recall and the precision can be improved by adjusting the various thresholds. For most of the time, the improvement of one comes at the expense of the other. For example, we can improve the recall by lowering $ID_T$ and broadening the range between $IDR^+$ and $IDR^-$ at the expense of precision. In other words, more pairs with ID values exceeding the threshold $ID_T$ and more pairs will fall within the range of acceptable IDR. In this case, we will lower the number of false negatives and hence, higher recall. Similarly, we can improve the precision by increasing $MI^+$. In this case, an increasing number of pairs will have mediocre mutual information (i.e. within the range $MI^+$ and $MI^-$). Consequently, the number of false positives will reduce when such pairs with mediocre mutual information are not merged due to their inability to satisfy additional constraints in the forms of $ID_T$, $IDR^+$ and $IDR^-$.

Due to the lack of existing dedicated techniques for measuring unithood, we were unable to perform a comparative study. Nonetheless, the high accuracy and F-score presented during our evaluation, and our analysis on the false positives and the false negatives revealed the potentials of our new measures in terms of high precision and recall, portability across domains, and configurability of the performance.

## 5   Applicability to Named-Entity Recognition

Named-entity recognition is one of the important tasks in information extraction. It involves the identification of noun phrases or more specifically, proper names from free text, and their classification into one of the many categories such as persons, geographical locations and companies. A typical named-entity recogniser performs part-of-speech tagging, and rely on patterns, heuristics and dictionary to identify proper names. The use of machine learning and other probabilistic methods such as support vector machines (Mayfield et al. (2003)) and hidden Markov models (Mittal et al. (1999)) have also gained popularity.

Most of these existing techniques for named-entity recognition works well when they are dealing with single-word names or sequences of nouns. In the face of more complex named-entities that consist of other part-of-speech especially prepositions, these techniques performed poorly. For example, most existing techniques would have to rely on heuristics and dictionaries to differentiate between *"Barrow in Furness"* and *"countries in Asia"*. Unfortunately, there are many problems related to the use of heuristics and dictionaries. For one, the maintenance of such dictionaries are costly and difficult. Questions about the reliability of stop-word boundaries, words capitalisation and punctuations arise regarding the use of heuristics and linguistics cues. For example, how can named-entity recognisers rely on capitalised words in the case of *"Barrow in Furness"* versus *"Perth in Australia"*? Besides prepositions, the conjunction *"and"* in named-entities posed similar challenge. According to Osenova and Kolkovska (2002), *"...problem arises when named-entity is a phrase, comprising a conjunction..."*. In such cases, the named-entity recogniser only recognises part of the named-entity. For example, in the case of *"Centre for Disease Control and Prevention"*, only *"Centre for Disease Control"* will be extracted. Other researchers (Mani et al. (1996)) have taken the default step of simply grouping name segments separated by prepositions or conjunctions into longer names.

Our new approach of deciding on whether two word sequences are to be merged or not is highly applicable to many areas in natural language processing especially named-entity recognition. The absence of any predefined resources in our approach will solve all the problems highlighted in the previous paragraph. Using our $UH(a_x, a_y)$ function, named-entity recogniser can easily determine whether or not parts of proper names should be merged together without ever relying on unreliable heuristics, and domain-restricted patterns and dictionaries.

## 6   Conclusion and Future Work

Many researchers inappropriately assume that termhood subsumes unithood. In this paper, we highlighted the significance of unithood and that its measurement should be given equal attention by researchers in automatic term recognition. The potential of unithood measurements can be extended to other areas in natural language processing such as noun-phrase chunking and named-entity recognition.

We proposed a new approach that provides dedicated measures specialised in measuring unithood. The first measure employs mutual information $MI(a_x, a_y)$ to capture the interdependence of the existence of $a_x$ and $a_y$, and allows us to determine if $a_x$ and $a_y$ are better off separated or otherwise in order to produce a stronger unit. The second measure, defined as *Independence (ID)* was inspired by *Cvalue*, and is meant to provide additional evidence in the determination of unithood. These two measures are combined into a Boolean function defined as *Unithood* ($UH$) that decides on whether $a_x$ and $a_y$ should be combined to form $s$.

Our evaluations revealed a precision and recall of 98.68% and 91.82% respectively with an accuracy at 95.42% in measuring the unithood of 1005 test cases. Due to the lack of existing dedicated techniques for measuring unithood, we were unable to perform a comparative study. Nonetheless, the excellent evaluation results together with the real-world text employed in our evaluation demonstrate the strengths of our new approach in regards to the determination of unithood. One of the future works that we plan to undertake is to increase the size of our test set to further establish the advantages of our new approach demonstrated through the current evaluations. We are planning to reduce the number of thresholds and at the same time, find ways to automatically optimise them.

**Acknowledgement**

This research was supported by the Australian Endeavour International Postgraduate Research Scholarship, and the Research Grant 2006 by the University of Western Australia.

**References**

D. Bourigault and C. Jacquemin. 1999. Term extraction + term clustering: An integrated platform for computer-aided terminology. In *Proceedings of the European Chapter of the Association for Computational Linguistics (EACL)*. Bergen.

K. Church and P. Hanks. 1990. Word association norms, mutual information, and lexicography. *Computational Linguistics*, 16(1):22–29.

I. Dagan and K. Church. 1994. Termight: Identifying and translating technical terminology. In *Proceedings of the 4th Conference on Applied Natural Language Processing*. Germany.

T. Dunning. 1994. Accurate methods for the statistics of surprise and coincidence. *Computational Linguistics*, 19(1):61–74.

K. Frantzi. 1997. Incorporating context information for the extraction of terms. In *Proceedings of the 35th Annual Meeting on Association for Computational Linguistics*. Spain.

K. Frantzi and S. Ananiadou. 1997. Automatic term recognition using contextual cues. In *Proceedings of the IJCAI Workshop on Multilinguality in Software Industry: the AI Contribution*. Japan.

A. Franz. 1997. Independence assumptions considered harmful. In *Proceedings of the 8th Conference on European Chapter of the Association for Computational Linguistics*. Madrid, Spain.

C. Kit. 2002. Corpus tools for retrieving and deriving termhood evidence. In *Proceedings of the 5th East Asia Forum of Terminology*. Haikou, China.

D. Klein and C. Manning. 2003. Accurate unlexicalized parsing. In *Proceedings of the 41st Meeting of the Association for Computational Linguistics*.

D. Kurz and F. Xu. 2002. Text mining for the extraction of domain relevant terms and term collocations. In *Proceedings of the International Workshop on Computational Approaches to Collocations*. Vienna.

I. Mani, R. Macmillan, S. Luperfoy, E. Lusher, and S. Laskowski. 1996. Identifying unknown proper names in newswire text. In *Proceedings of the ACL SIG Workshop on Acquisition of Lexical Knowledge from Text*.

J. Mayfield, P. McNamee, and C. Piatko. 2003. Named entity recognition using hundreds of thousands of features. In *Proceedings of the Conference on Computational Natural Language Learning*.

V. Mittal, S. Baluja, and R. Sukthankar. 1999. Applying machine learning for high-performance named-entity extraction. In *Proceedings of the Conference of the Pacific Association for Computational Linguistics*.

H. Nakagawa and T. Mori. 2002. A simple but powerful automatic term extraction method. In *Proceedings of the International Conference On Computational Linguistics (COLING)*.

P. Osenova and S. Kolkovska. 2002. Combining the named-entity recognition task and np chunking strategy for robust pre-p. In *Proceedings of the Workshop on Linguistic Theories and Treebanks*. Sozopol, Bulgaria.

W. Wong. 2005. *Practical Approach to Knowledge-based Question Answering with Natural Language Understanding and Advanced Reasoning*. Master's thesis, National Technical University College of Malaysia.